\documentclass[english,]{article}
\usepackage{lmodern}
\usepackage{amssymb,amsmath}
\usepackage{ifxetex,ifluatex}
\usepackage{fixltx2e} 
\ifnum 0\ifxetex 1\fi\ifluatex 1\fi=0 
  \usepackage[T1]{fontenc}
  \usepackage[utf8]{inputenc}
\else 
  \ifxetex
    \usepackage{mathspec}
  \else
    \usepackage{fontspec}
  \fi
  \defaultfontfeatures{Ligatures=TeX,Scale=MatchLowercase}
\fi
\IfFileExists{upquote.sty}{\usepackage{upquote}}{}
\IfFileExists{microtype.sty}{%
\usepackage[]{microtype}
\UseMicrotypeSet[protrusion]{basicmath} 
}{}
\PassOptionsToPackage{hyphens}{url} 
\usepackage[unicode=true]{hyperref}
\hypersetup{
            pdftitle={Mixing syntagmatic and paradigmatic information for concept detection},
            pdfauthor={Louis Chartrand and Mohamed Bouguessa},
            pdfborder={0 0 0},
            breaklinks=true}
\ifnum 0\ifxetex 1\fi\ifluatex 1\fi=0 
  \usepackage[shorthands=off,main=english]{babel}
\else
  \usepackage{polyglossia}
  \setmainlanguage[]{english}
\fi
\usepackage{longtable,booktabs}
\IfFileExists{footnote.sty}{\usepackage{footnote}\makesavenoteenv{long table}}{}
\usepackage{graphicx,grffile}
\makeatletter
\def\maxwidth{\ifdim\Gin@nat@width>\linewidth\linewidth\else\Gin@nat@width\fi}
\def\maxheight{\ifdim\Gin@nat@height>\textheight\textheight\else\Gin@nat@height\fi}
\makeatother
\setkeys{Gin}{width=\maxwidth,height=\maxheight,keepaspectratio}
\IfFileExists{parskip.sty}{%
\usepackage{parskip}
}{
\setlength{\parindent}{0pt}
\setlength{\parskip}{6pt plus 2pt minus 1pt}
}
\providecommand{\tightlist}{%
  \setlength{\itemsep}{0pt}\setlength{\parskip}{0pt}}
\ifx\paragraph\undefined\else
\let\oldparagraph\paragraph
\renewcommand{\paragraph}[1]{\oldparagraph{#1}\mbox{}}
\fi
\ifx\subparagraph\undefined\else
\let\oldsubparagraph\subparagraph
\renewcommand{\subparagraph}[1]{\oldsubparagraph{#1}\mbox{}}
\fi

\makeatletter
\def\fps@figure{htbp}
\makeatother

\makeatletter
\@ifpackageloaded{subfig}{}{\usepackage{subfig}}
\@ifpackageloaded{caption}{}{\usepackage{caption}}
\AtBeginDocument{%

}
\AtBeginDocument{%

}
\@ifpackageloaded{float}{}{\usepackage{float}}
\floatstyle{ruled}
\@ifundefined{c@chapter}{\newfloat{codelisting}{h}{lop}}{\newfloat{codelisting}{h}{lop}[chapter]}
\floatname{codelisting}{Listing}

\makeatother

\title{Mixing syntagmatic and paradigmatic information for concept detection}
\author{Louis Chartrand and Mohamed Bouguessa}
\date{}

\begin{document}
\maketitle
\begin{abstract}
In the last decades, philosophers have begun using empirical data for
conceptual analysis, but corpus-based conceptual analysis has so far
failed to develop, in part because of the absence of reliable methods to
automatically detect concepts in textual data. Previous attempts have
shown that topic models can constitute efficient concept detection
heuristics, but while they leverage the syntagmatic relations in a
corpus, they fail to exploit paradigmatic relations, and thus probably
fail to model concepts accurately. In this article, we show that using a
topic model that models concepts on a space of word embeddings (Hu and
Tsujii, 2016) can lead to significant increases in concept detection
performance, as well as enable the target concept to be expressed in
more flexible ways using word vectors.
\end{abstract}

\section{Introduction}\label{sec:lctmintro}

Conceptual analysis has, in one form or another, long been a staple of
philosophical methodology (Beaney 2018). When considered as a method,
conceptual analysis often requires the input of empirical data, be it in
the form of perceptual data, like philosophical intuitions (Pust 2000),
or in other measurable forms. In the last decades, philosophers have
begun performing experiments to get a grasp of human's conceptual
behaviour and reactions in order to get a more precise understanding of
fundamental concepts like KNOWLEDGE, JUSTICE or RESPONSIBILITY. However,
these methods are limited, as controlling for variables forces
experimenters to put participants in somewhat unnatural situations and
create studies with low ecological validity.

Other voices have suggested that concepts could fruitfully be studied in
textual corpora (Meunier, Biskri, and Forest 2005; Bluhm 2013; Andow
2016; Chartrand 2017). They argue that methods based on the
distributional hypothesis, and that hail from subfields such as natural
language processing, text mining and corpus linguistics, could shed
light on at least some of the concepts that are objects of philosophical
scrutiny.

However, these methods are not well-tuned to philosophical conceptual
analysis, as they usually rely heavily on keywords as an indicator of
the presence of a concept. While concepts are often associated to
words---indeed, even in philosophical discussions, we usually use words
as tags for concepts---words are usually associated to more than one
concept, and the one that is expressed in any instance is determined by
the context. Conversely, concepts can be expressed in a variety of ways.
Not only can two or more words or word compounds express the same
concept, but concepts don't always attach to words: they can be present
in a sentence thematically, or because of an inference that one must
make when decoding the sentence. For conceptual analysts who would like
their idea of a concept to be representative of all its various uses,
this can be an important problem.

The first problem can often be circumvented with a judicious choice of
corpus, as controlling the context can often control the sense a word
will espouse. As a result, recent work on concept detection has focused
on the second problem in a bid to detect where a concept is expressed
without the word it is most associated with (Pulizzotto et al. 2016;
Chartrand et al. 2016; Chartrand, Cheung, and Bouguessa 2017). While
these efforts have yielded promising results, these methods still rely
on the identification of a concept with a word in order both for
modelling the constitution of higher-level discursive entities (like
topics or narratives) and for representing the queried concept.

This assumption is problematic in at least two ways. On the one hand,
these higher-level entities are not composed of words, but of concepts.
There is thus the worry that representing them as composed of words
makes for an imprecise model. Another concern is with the word-sense
ambiguity: a single word may refer to two different concepts, depending
on its pragmatic and textual context. One way textual data analysts have
dealt with this problem has been to tailor corpora to fit their needs,
and choose corpora where the concepts they are interested in happen to
be unambiguously associated to a word or a word expression.

Our hypothesis is that these two concerns can be addressed by
representing concepts not as words or word expressions, but coordinates
on a word embedding space---or, to be more precise, \(n\)-dimensional
vectors whose semantic properties are determined by their distances to
other \(n\)-dimensional vectors that represent words ands concepts from
a text corpus. In other words: representing concepts as such enable us
(1) to make a better model of higher-level entities like topics, which
in turn translates in better performances in concept detection and (2)
to formulate queries when concepts do not perfectly match with a word or
word expression in the corpus.

To test this hypothesis, we employ Hu's and Tsujii's (2016) Latent
Concept Topic Model (LCTM) to construct processing chains for concept
detection. The LCTM constructs topics as distributions over concepts,
which are coordinates in a word embedding space, making for a model that
is more theoretically coherent with (Chartrand no date) than previous
concept detection models. The processing chains can then be applied to a
corpus (in this case, decisions from the Québec Court of Appeal, as in
Chartrand, Cheung, and Bouguessa 2017) and tested against human
annotations.

In section~\ref{sec:prevWork}, we review the relevant litterature. We
give an overview of the state of the art in topic modelling and word
embeddings, and then review hybrid models. In section
section~\ref{sec:task}, we formulate the concept detection task and
explain how the annotations forming the gold standard are gathered. In
section~\ref{sec:models}, we describe the underlying model and
functionning of LCTM, and how it links with the theory that underlies
the concept detection task. In sections \ref{sec:method} and
\ref{sec:experimentation}, the experiments and their application are
described, and in section~\ref{sec:results}, we review the results,
which are discussed in section section~\ref{sec:discussion}.

\subsection{Previous work}\label{sec:prevWork}

From the beginning, topic models have tried to model concepts as an
underlying dimension of the text: latent semantic indexing (Deerwester
et al. 1990) described documents in terms of latent ``concepts''.
However, by the end of the 90s, Hofmann (1999) described the latent
variables in his probabilistic latent semantic indexing as ``class
variables'', and Blei, Ng, and Jordan (2003) called them ``topics'' in
his latent Dirichlet allocation (LDA) model. Gabrilovich and Markovitch
(2007) resurrected the idea of a latent semantic dimension as a concept
by forming representations from Wikipedia articles, but their reported
success seems to hail from a mere size effect rather than Wikipedia's
grouping of discourse under labels (Gottron, Anderka, and Stein 2011).

While there are numerous variations, the topic models that are
well-known in the natural language processing community treat topics as
a mixture of variables that are of the same kind, that influence word
occurrences in the same way, and that thus have \emph{a priori} the same
role in shaping discourse. Once learned, they capture the
syntagmatic\footnote{For a more thorough account of syntagmatic and
  paradigmatic relations in the context of distributional semantics, cf.
  Sahlgren (2008). Cf. also pages
  \pageref{defn:syntagparad}-\pageref{end:defn:syntagparad} of the
  present thesis.} relations between words. Words are syntagmatically
close when they participate in the same discourse units---in other
words, when they are neighbours, or when they tend to come together.

Word embeddings, on the other hand, tend to capture paradigmatic
relations. Words are
paradigmatically\footnotemark[\value{footnote}] close when they tend to
have the same neighbours. As a result, they tend to have similar roles
in discourse, and therefore be synonyms or antonyms\footnote{Antonyms
  are nearly identical except on one semantic dimension, on which they
  are opposites. This is why they typically have very similar roles in
  discourse and sentences.}. Word embeddings evolved from language
models in the early 2000s (Bengio et al. 2003), but were then too
computationally expensive to be applied to large corpora. Collobert and
Weston (2008), followed by Mikolov and colleagues (Mikolov, Chen, et al.
2013; Mikolov, Sutskever, et al. 2013), found ways to get the computing
cost down, opening the way for word embeddings to become an essential
part of the natural language processing toolkit.

Given the popularity of topic models and the word embedding boom that
followed word2vec (Mikolov, Chen, et al. 2013), it is no wonder that
many attempts to combine them have been made. Several of them (Nguyen et
al. 2015; C. Li et al. 2016; Hu and Tsujii 2016; Wang et al. 2017; Le
and Lauw 2017; Li et al. 2018; Peng et al. 2018; Zhang, Feng, and Liang
2019) aim at making topic models that work well with short documents
like tweets, where too few words are employed (sparsity problem). Others
target the problem of homonymy/polysemy (Liu et al. 2015; Law et al.
2017), seek more interpretable topics (Potapenko, Popov, and Vorontsov
2017; Zhao et al. 2018), or aim at exploiting complementary
representations (S. Li et al. 2016; Moody 2016; Bunk and Krestel 2018).
Often word embeddings are simply seen as a means to make a more
realistic model (Das, Zaheer, and Dyer 2015; Batmanghelich et al. 2016;
Hu and Tsujii 2016; X. Li et al. 2016; Xun et al. 2017).

For example, there is a lineage of models that can be seen as attempts
to see how word embeddings fit in the generative story behind
probabilistic topic models. Das \emph{et al.}'s Gaussian LDA (2015)
replaces the word-over-topic distribution of the LDA with coordinates on
the word embedding space. A word's probability given a topic associated
with such coordinates are then inferred from the corresponding word
embedding's proximity using the Gaussian distribution. Batmanghelich et
al. (2016) starts from the Gaussian LDA and replaces the Gaussian
distribution with the von Mises-Fisher (vMF) distribution, which is a
probability distribution over angles centered on a vector. Hu and Tsujii
(2016) and X. Li et al. (2016) both choose not to identify topics to
coordinates or vectors on the word embeddings space, but rather model
topics as constituted by such objects. In the former, topics are
distributed over these objects (which are called ``concepts''), and word
probabilities are inferred from concepts using a Gaussian distribution,
while the latter identifies topics as complex von Mises-Fisher mixtures
over a determined number of bases. Bunk and Krestel (2018) go for a
middle-ground position, where words are both influenced by typical
LDA-style topics and GLDA-style vector-topics that are situated in the
word embeddings space. Perhaps more interestingly, they report no
advantage in using mixture models or vMF distributions over simple
GLDA-style gaussian distributions, at least in terms of topic coherence
and word intrusion tasks.

Perhaps because it is specifically tailored for the needs of
philosophical conceptual analysis, few attempts have been made so far at
addressing the concept detection task (Chartrand et al. 2016; Pulizzotto
et al. 2016; Chartrand, Cheung, and Bouguessa 2017). These papers
emphasize the inadequacy of using keywords to recall text segments where
a concept is expressed, but their models still use words as stand-ins
for concepts both for articulating queries and for modelling the
concept-topic relationship.

This inadequacy calls for alternate models. However, given the large
variety of existing topic models, we might not have to create a new one.
Chartrand (no date) argues that higher-level discourse entities (which
can arguably be modelled by LDA-style topics) are constructed from
concepts rather than words, topic models that use the word embedding
space to model concepts over which topics are distributed. This suggests
that topics models where topics are distributed over concepts (rather
than words) in the word embedding space (Hu and Tsujii 2016; X. Li et
al. 2016) are more likely to accurately represent topics and their
structure. One can hope that such a representation of concepts and their
association with topics will yield better results on the concept
detection task.

\section{The concept detection task}\label{sec:task}

As Haslanger (2012) argues, philosophical conceptual analysis can pursue
different aims. In some cases, the goal is to represent the concept that
we (collectively) have as we possess it. We can call this a
\emph{conceptualistic} conceptual analysis. In other cases, the idea is
to represent the concept as it functions: this would be a
\emph{functionalistic} conceptual analysis. In the case of a concept
that represents something, this means that our objective here is to
represent the concept so as to reflect its referent rather than our
common account of it. For instance, if the function of the concept is to
refer to dolphins, then it would not matter if most of us thought of
dolphins as fish or if we thought that they have wings: a proper
functionalist analysis would still represent dolphins as wingless
aquatic mammals. Finally, we make an \emph{ameliorative} conceptual
analysis when the goal of the analysis is to produce a concept that
better fulfills the role it plays in discourse, knowledge or society.

This diversity in purpose, however, branches from common grounds.
Firstly, there is a sense in which conceptual analysis always is
ameliorative, as the representation it aims to make is itself a new
concept, meant to play (most often) new roles, if only in philosophical
conversations. Secondly, no matter our purpose, it ought to start with
an understanding of how this concept functions in its community's
discourses and ways of life. Therefore, conceptual analysis demands a
thorough picture of a concept's usage, which is where natural language
processing can lend a hand.

This thorough picture demands that we be able to capture a concept in as
large a variety of uses as possible. While NLP can only be of help when
it comes to observing discourse, it is important to include, as much as
possible, all ways by which a concept is employed in discourse. As
argued by Chartrand (no date), much like words bind together to form
sentences, concepts bind together to form higher-level entities that are
reproduced in a community. These entities can be themes, narratives,
arguments, etc. To understand such a higher-level entity, one needs to
understand all of its components---therefore, a concept is always
present when a topic or a narrative is expressed. However, a concept
might not be present in the form of the word that, in proper context, we
most readily associate with it (say the word ``dolphin'' for the concept
DOLPHIN). It might present itself in the form of an anaphor (``it'',
``them''), an hypernym (``the animal'') or a description (``these
long-nosed swimmers''). It might also be implicitly present within a
hidden premise, as part of a piece of information that can be inferred
from the text, as part of the background knowledge that we access in
order to understand what is being communicated, or even as the object
about which we are implicitly talking about. Being present in different
ways in discourse often means that a concept is employed differently
and, therefore, has different roles. As a result, it is important that a
concept detection algorithm be able to capture the different ways a
concept is present in the text.

Concept detection is thus distinct from more traditional information
retrieval problems: here, it is not relevance that is sought, but
presence. The challenge is not to find the most relevant passages for
the expression of a concept, but to find all text segments where is it
present. It is also different from such problems as word-sense
extraction or ontology learning because concepts need not be associated
with words.

\section{Models}\label{sec:models}

As mentioned in section~\ref{sec:prevWork}, there are considerations
which lead us to hypothesize that, for concept detection, models that
represent concepts in their generative story are more likely to reflect
the topic structure in such a way that it can be leveraged for concept
detection. Perhaps more interestingly, explicit modelling of concepts
(as opposed to simply leveraging word embedding data to direct the
learning process) makes it possible to formulate queries using word
combinations, which can help disambiguate the query (e.g.
\(bank - river\) might yield the concept of BANK as this place where we
make financial transaction) or make it possible to look for new
concepts.\footnote{This is also why topic models based on word
  embeddings are, in this context, a superior solution to algorithms
  that use concept databases, such as Tang et al. (2018), or algorithms
  that model concepts simply as latent variables, as El-Arini, Fox, and
  Guestrin (2012).}

This leaves us with the LCTM (Latent Concept Topic Model, Hu and Tsujii
2016) and MvTM (Mix von Mises-Fisher Topic Model, X. Li et al. 2016)
models. However, the MvTM makes counter-intuitive assumptions concerning
the availability of concepts for constituting topics. There are two
variants to the MvTM: the ``disjoint bases'' variant
(MvTM\textsubscript{d}), in which topic mixtures are made from bases
that cannot be shared with other topics, and the ``overlapping bases''
variant (MvTM\textsubscript{o}) where topic mixtures are partly made
from bases that can be shared with other topics. If bases were meant to
model concepts, then we would expect all of them to be shared by many
topics. According to the authors, this serves to prevent identical
topics from emerging, but language is too fluid to afford such a
restriction: no theme, narrative or argument ever has had exclusive
rights to a concept. As a result, LCTM seems like the better
alternative.

The LCTM is an evolution of the LDA (Blei, Ng, and Jordan 2003) and GLDA
(Das, Zaheer, and Dyer 2015) models, all three of which are
probabilistic graphical models. This is to say that they rely on a
generative model, which represents an abstract hypothesis of how a text
is constructed and structured.

\subsection{LDA}\label{sec:ldamodel}

In the LDA model, topics are represented by two variables: a multinomial
distribution over documents (\(\theta\)), and multinomial distribution
over words (\(\phi\)). These distributions are designed to be sampled
from the conjugate Dirichlet priors with parameters \(\alpha\) and
\(\beta\) respectively. In Blei's (Blei, Ng, and Jordan 2003) account,
model uses this generative process, which assumes a corpus \(D\) of
\(M\) documents each of length \(N_{i}\):

\begin{itemize}
\tightlist
\item
  Draw \(\theta_{i}\sim\mathrm{Dirichlet}\left(\alpha\right)\), where
  \(i\in\left\{ 1\ldots M\right\}\), the topic distribution for document
  \(i\)
\item
  Draw \(\phi_{k}\sim\mathrm{Dirichlet}\left(\beta\right)\), where
  \(k\in\left\{ 1\ldots K\right\}\), the word distribution for topic
  \(k\)
\item
  For each of the word positions \(i,j\), where
  \(j\in\{1,\dots,N_{i}\}\), and \(i\in\{1,\dots,M\}\):
\item
  Draw a topic \(z_{i,j}\sim\mathrm{Multinomial}(\theta_{i})\).
\item
  Draw a word \(w_{i,j}\sim\mathrm{Multinomial}(\varphi_{z_{i,j}}).\)
\end{itemize}

\subsection{GLDA}\label{sec:gldamodel}

With their GLDA model, Das, Zaheer, and Dyer (2015) replace \(\phi_k\)
with a covariance \(\Sigma_k\) and coordinates to a point that acts as
its distribution's mean \(\mu_k\). The covariance \(\Sigma_k\) is
sampled from an inverse Wishart distribution, and the mean \(\mu_k\) is
sampled from a normal distribution centered at zero (\(\mu\)). Thus,
GLDA's generative story goes like this:

\begin{itemize}
\tightlist
\item
  For each topic \(k \in \{1\dots K\}\)
\item
  Draw a topic covariance
  \(\Sigma_k \sim \mathcal{W}^{-1}({\mathbf\Psi},\nu)\)
\item
  Draw a topic mean
  \(\mu_k \sim \mathcal N(\mu, \frac{1}{\kappa}\Sigma_k)\)
\item
  For each document \(i \in \{1\dots M\}\)
\item
  Draw a topic distribution \(\theta_i \sim \mathrm{Dirichlet}(\alpha)\)
\item
  For each word \(w \in \{1\dots N_i\}\)

  \begin{itemize}
  \tightlist
  \item
    Draw a topic \(z_w \sim \mathrm{Multinomial}(\theta_i)\)
  \item
    Draw a word vector \(v_w \sim \mathcal N(\mu_{z_w},\Sigma_{z_w})\)
    (the chosen word is the one whose word embedding is closest to
    \(v_w\))
  \end{itemize}
\end{itemize}

\subsection{Word embeddings}\label{sec:wemodel}

Word embeddings have developed as a way of representing the semantic
information of words in a corpus (paradigmatic relations in particular),
and they are employed as such in the LCTM model.

Word embeddings tap in the power of term-term cooccurrence vectors. A
term-term cooccurrence matrix is a \(N \times N\) matrix \(M\), where
\(N\) is the number of word types in a corpus, and where the value of
each cell \(w_{i,j}\) is equal to the number of times the
\(i^{\textrm{th}}\) and \(j^{\textrm{th}}\) cooccur within a window of
\(k\) words. A cooccurrence vector \(v_i\) is the \(i^{\textrm{th}}\)
row of matrix \(M\) and corresponds to the \(i^{\textrm{th}}\) word.
Cooccurrence vectors whose cosine distance are small are typically
semantically close in the sense that they are often synonyms or
antonyms. In other words, they are paradigmatically related.

Because term-term cooccurrence vectors tend to be very large, especially
in big corpora, there is an incentive to compress them to make them more
manageable through dimensionality reduction. Thus, words are associated
with a \(k\)-dimensional vector, where \(k\) is an arbitrary number,
typically between 50 and 300.

Dimensionality reduction can be achieved by many means. So-called
``count'' methods (Baroni, Dinu, and Kruszewski 2014; Pennington,
Socher, and Manning 2014) use methods such as singular value
decomposition and matrix factorization to reduce weighted count vectors
(weighting schemes include positive pointwise mutual information and
local mututal information). Meanwhile ``predict'' methods (Bengio et al.
2003; Collobert and Weston 2008; Mikolov, Chen, et al. 2013; Mikolov,
Sutskever, et al. 2013) set up neural networks that simultaneously learn
to predict a word from its a small context window\footnote{Classically,
  this meant a small window before the target word (Bengio et al. 2003;
  Collobert and Weston 2008) (this would be the classic ``language
  model'' paradigm), but Mikolov, Chen, et al. (2013) have introduced
  the Skip-gram model, where a word is used to predict the words
  immediately before and after it, within a small window, and the CBOW
  model, where the context words are used to predict the target word.
  These models have since then become the norm.} and learn vector
representations for each word type.

\subsection{LCTM}\label{sec:lctmmodel}

With the LCTM, Hu and Tsujii (2016) act on the intuition that topics do
not model the same kind of distributional similarity that are modeled
with word embeddings. As they note, words that are topically close, like
``neural'' and ``network'' in a computer science corpus, will be far
away on a word embedding space. This is why they see topics as
distributed over other latent variables which they call
\emph{concepts}\footnote{While here ``concept'' is a technical term that
  refers to features of the LCTM, we believe this use is justified as
  this concept of CONCEPT can be argued to be an \emph{explication} (in
  Carnap's (1950) sense) of the concept of CONCEPT that is defended in
  Chartrand (no date). In other words, for the purpose of building an
  algorithm, it is a more precise, more explicit version of the latter
  concept, that retains some of its features and enables us to say
  something about the former. In particular, we assume that instantions
  of the technical sense of CONCEPT can tell us something about where,
  in the text corpus, corresponding instantiations of the non-technical
  sense of CONCEPT are mobilized. For example, if a technical concept,
  represented as coordinates in a word embedding space, has for
  corresponding lay concept the concept that we associate with the word
  ``dolphin'' (therefore, the lay concept DOLPHIN), then we expect this
  technical concept to help us determine where the lay concept DOLPHIN
  is mobilized.}, and which are represented by coordinates in the word
embedding space. In other words, if a concept is active at a certain
point in the text, then the words whose word embeddings are close to the
concept's are more likely to appear there.

The LCTM's generative model goes like this, with \(C\) being the number
of concepts in the model:

\begin{itemize}
\tightlist
\item
  For each topic \(k \in \{1\dots K\}\)
\item
  Draw a topic concept distribution
  \(\phi_k \sim \mathrm{Dirichlet}(\beta)\)
\item
  For each concept \(c \in \{1\dots C\}\)
\item
  Draw a concept vector
  \(\mu_c \sim \mathcal N (\mu, \sigma^2_0 \pmb I)\)
\item
  For each document \(i \in \{1\dots M\}\)
\item
  Draw a topic distribution \(\theta_i \sim \mathrm{Dirichlet}(\alpha)\)
\item
  For each word \(w \in {1\dots N_i}\)

  \begin{itemize}
  \tightlist
  \item
    Draw a topic \(z_w \sim \mathrm{Multinomial}(\theta_i)\)
  \item
    Draw a concept \(c_w \sim \mathrm{Multinomial}(\phi_{z_w})\)
  \item
    Draw a word vector \(v_w \sim \mathcal N(\mu_{c_w},\Sigma_{z_w})\)
    (the chosen word is the one whose word embedding is closest to
    \(v_w\))
  \end{itemize}
\end{itemize}

\begin{figure}
\centering
\includegraphics{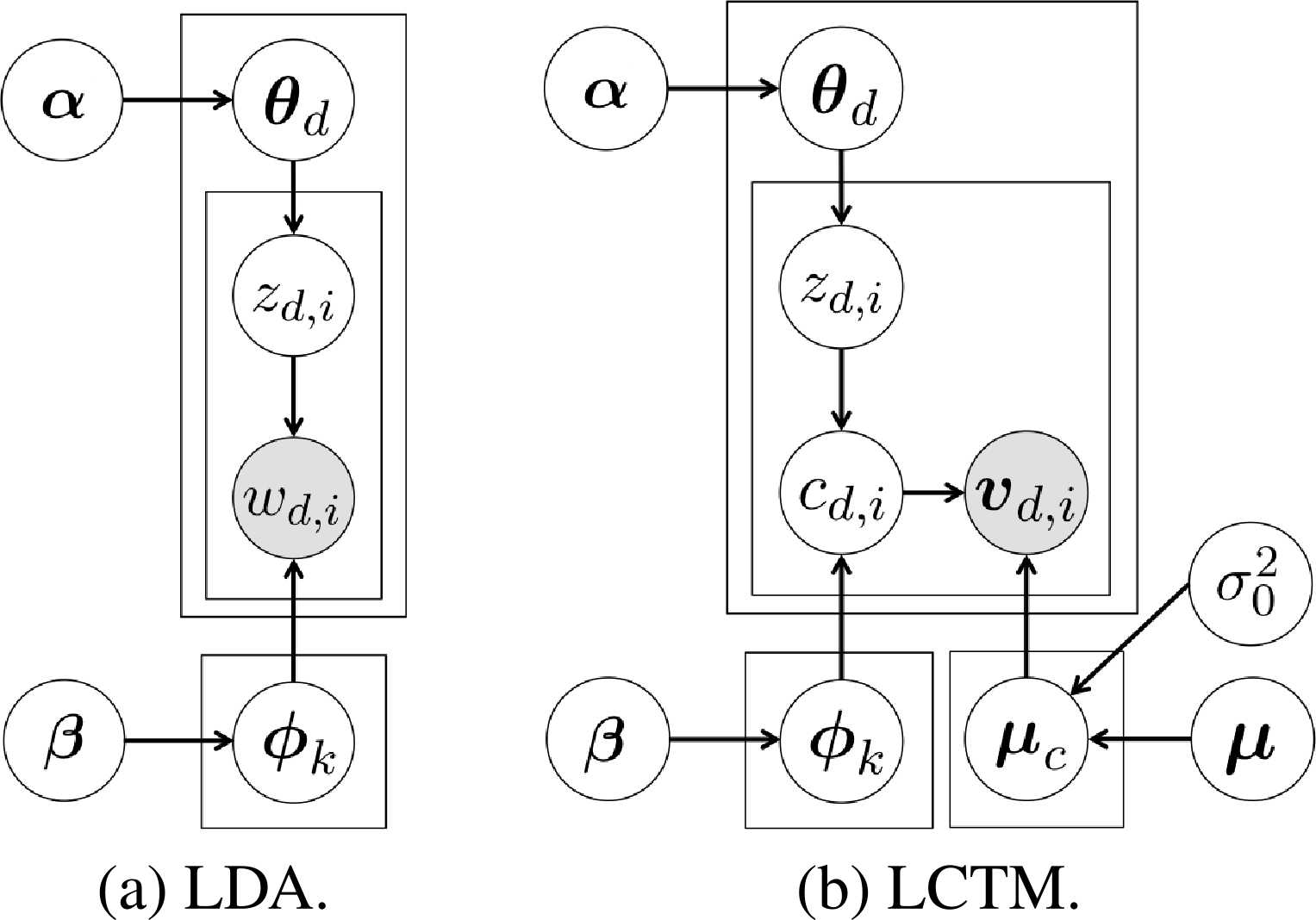}
\caption{Plate models for LDA and LCTM}\label{fig:platemodels}
\end{figure}

The graphical model for LDA and LCTM are shown in
figure~\ref{fig:platemodels}.

In relation with the problem of word-sense ambiguity mentioned in
section section~\ref{sec:lctmintro}, it is interesting to note that the
same word can be highly likely for different concepts belonging to
different topics, which themselves are associated with whole documents.
As such, the context of the document determines the topics, and
therefore the concept to which a word can be associated. In so doing,
the LCTM model can associate different concepts to the same word type
depending on the context of the document it is in, and can thus
disambiguate between different senses of a word.

\section{Method}\label{sec:method}

\subsection{Inference}\label{inference}

Given its relative simplicity and its efficiency, Gibbs sampling is by
far the most popular method for learning the parameters of topic models
that employ word embeddings. LCTM is no exception.

During the inference process, both concept and topic assignments for
each word are sampled, using those two equations:

\begin{equation} p\left(z_w=k\middle|c_w=c,\pmb z^{-w},\pmb c^{-w}, \pmb v\right) \propto \left(n^{-w}_{i,k}+\alpha_k\right) \cdot \frac{n^{-w}_{k,c}+\beta_c}{n^{-w}_{k,\cdot}+\sum_{c^\prime \in \left\{1\dots C\right\}}\beta_{c^\prime}} \label{eq:samp1}\end{equation}

\begin{equation} p\left(c_w=c\middle|z_w=k,\pmb z^{-w},\pmb c^{-w}, \pmb v\right) \propto \left(n^{-w}_{c,k}+\beta_c\right)\cdot \mathcal N \left(v_w\middle|\overline \mu_c\,\sigma_c^2\pmb I\right) \label{eq:samp2}\end{equation}

\subsection{Concept extension}\label{concept-extension}

\begin{figure}
\centering
\includegraphics[width=5.30000in,height=3.00000in]{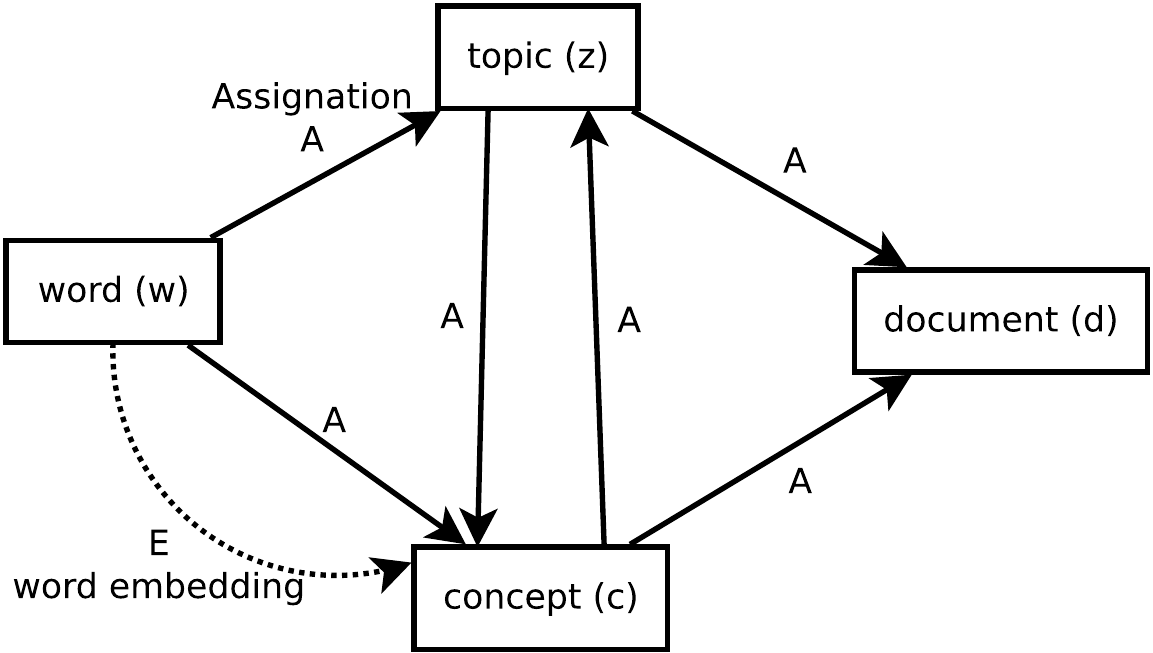}
\caption{Constructing a concept extension chain from an LCTM
model.}\label{fig:cechain}
\end{figure}

Once the model has been learned, we have, for each word position,
assignment to a concept and a topic, on top of information about its
word type and document membership that was provided to the LCTM.
Furthermore, we have vectors for each concept and we had provided a
vector for each word, all in the same word embeddings space.

Concept detection formally consists in a function that yields a set of
documents from a query, which can be either a word type or a vector in
the word embeddings space. From the information LCTM produces, there are
a number of ways we could make such a function. For instance, we could
find the concepts assigned to the query word, and then find all the
documents where these concepts are assigned. But we could also find the
word vector for this word, then retrieve the closest concept (in terms
of cosine similarity), find the topics where it is assigned, then find
the documents where these topics are assigned.

In order to represent the variety of ways a concept extension can be
obtained, we use a special notation (cf. figure~\ref{fig:cechain}).
``w'', ``c'', ``z'' and ``d'' respectively mean ``word'', ``concept'',
``topic'' and ``document''. Furthermore, ``q'' represents a query
expressed in the form of a vector. Transitions are noted ``E'' or ``A'':
``xEy'' means ``get the y whose vector is closest to x's vector'' and
``xAy'' means ``get all the ys which are assigned to a word where x is
also assigned''. Thus, the first example of the previous paragraph would
be noted ``wAcAd'' and the second example would be ``wEcAzAd''. Given
the nature of concept detection, the first letter is always either ``w''
or ``q'', and the last is always ``d''.

While the number of possible ways we can get a concept extension using
data from a LCTM model is potentially infinite, it makes no sense
looping over concepts and topics. Therefore, given a word query, only 8
variations are possible: ``wAcAd'', ``wAcAzAcAd'', ``wAcAzAd'',
``wAzAcAd'', ``wAzAd'', ``wEcAd'', ``wEcAzAcAd'' and ``wEcAzAd''. Given
a word vector query, only three variations are available: ``qEcAd'',
``qEcAzAcAd'' and ``qEcAzAd''.

\section{Experimentation}\label{sec:experimentation}

\subsection{Experiment 1}\label{experiment-1}

The first part of our research hypothesis stated that modelling concepts
in a topic model would lead to a generally better modelling of the text
structure. This, in turn, should lead to better concept detection
performance.

To evaluate this proposition, we test all 8 methods for concept
detection using LCTM as described in section~\ref{sec:method}. For
comparison, we are also testing the ``Online LDA-Top 30'' and the
``Gibbs sampling-Concrete Assignment'' heuristics from Chartrand,
Cheung, and Bouguessa (2017)\footnote{In this heuristic, words are
  associated to a topic if they are among the 30 words most likely to
  come up if this topic is activated. From this, we get the concept
  extension by recalling all the segments or documents in which any of
  the topics associated with the query word are activated.}, along with
the keyword heuristic (recall all text segments where the query word is
present). In this evaluation, concept extensions for 754 queries are
computed and evaluated against gold standards using the Matthew
correlation coefficient (MCC)\footnote{Here, IR standard metrics for
  evaluation like accuracy and F1-measure are not employed because our
  gold standard represents an unsual set, where annotated
  concept-segment pairs are much more likely to be positive than
  randomly chosen concept-segment pairs. Cf.
  section~\ref{sec:corpusannot}.}. All of these queries are formulated
as a single word; when computing an extension from a chain that begins
with ``wE'', we employ the word embedding corresponding to the query
word.

\subsection{Experiment 2}\label{experiment-2}

The second part of our research hypothesis suggested that using a topic
model like LCTM, that models concepts on the word embedding space, would
allow us to formulate queries for concepts that are not adequately
represented with a single word. To test this, we test 588 multiword
expressions and represent them on the word embedding space. To do this,
we exploit the compositional property of word embeddings, and represent
these expressions as the sum of the vectors corresponding to the content
words in the expression\footnote{Unlike in English, where compound words
  are created merely by putting the words together, compound words in
  French often involve prepositions that further constrains how the
  semantic composition should be interpreted. For simplicity's sake, we
  ignore this information here.}. Because LCTM makes no assignation for
multiword expressions, only the chains built for word vector
queries---those beginning with ``qE''---are available. They are compared
with the keyword heuristic against a gold standard using the MCC.

\subsection{Corpus \& pretreatment}\label{corpus-pretreatment}

Our corpus is composed of 186,860 segments extracted from 5,229
French-language court decisions of the Quebec Court of Appeal. These
decisions where all published between January 1, 2004 and December 31,
2014. Each segment corresponds to a numbered paragraph in this
judgements, and each is parsed with a POS tagger\footnote{\url{http://www.cis.uni-muenchen.de/~schmid/tools/TreeTagger/}}.
Only verbs, nouns, adverbs and adjectives are kept. Furthermore, judges
like to cite law articles, jurisprudence or doctrine in their
judgements; these citations have been removed.

Prior to applying the concept detection chains, a word embedding matrix
has been learned using gensim's implementation of word2vec\footnote{https://radimrehurek.com/gensim/}.
Vector size has been set to 100. The LCTM model was learned using Hu and
Tsujii's implementation,\footnote{https://github.com/weihua916/LCTM}
with the number of topics being 150 (same as in Chartrand, Cheung, and
Bouguessa 2017) and the number of concepts being 1,000. Two LDA models
were used for comparison: one is learned using Hoffman's Online LDA
algorithm (2010), as implemented in gensim\footnote{https://radimrehurek.com/gensim/};
the other is learned using Griffiths' and Steyvers' collapsed Gibbs
sampling (Griffiths and Steyvers 2004), as implemented by Blondel
(2010)\footnote{Code available at
  \url{https://gist.github.com/mblondel/542786}.}.

\subsection{Corpus annotations}\label{sec:corpusannot}

A subset of our corpus was annotated using the same two-step method
presented in Chartrand, Cheung, and Bouguessa (2017):

\begin{enumerate}
\def\labelenumi{\arabic{enumi}.}
\tightlist
\item
  Annotators were asked to read a text segment from the corpus, and then
  write down five concepts that were present in it---or, in other words,
  that contributed to what was being said.
\item
  Drawing from concepts obtained in step 1, segments were paired with
  six concepts. Annotators were asked to rate the concept's presence
  from 1 to 4, 1 being completely absent and 4 being highly present. For
  our purpose, we consider that a concept is present if the annotator
  scores more than 1---the scale was employed to avoid a concept to be
  tagged as absent if it was weakly or very implicitly present. The draw
  was tweaked so that, on average, annotators would generally be
  compelled to say that any given concept was present (2-4) roughly half
  of the time---this was to ensure that annotators would not be tempted
  to force a concept upon a segment to compensate for the fact that very
  little concept were marked as absent.
\end{enumerate}

These annotations were done, on the one hand, by domain
experts---jurists---and on the other hand by workers on the
crowdsourcing site \emph{Crowdflower} (which rechristened itself
\emph{Figure 8} before the end of the study). The 9 expert jurists
annotated 103 segments with 1,031 annotations. As for lay workers'
annotations, after screening for ``spam'' annotations (annotations that
seemed random or did not seem to reflect an actual understanding), we
were left with 3,240 annotations from 25 workers on 105 segments. While
there was little overlap in the segment-concept pairs annotated by
experts and non-experts, annotation patterns on the 130 pairs where
there was overlap reveal that annotations of experts and non-experts
correlate moderately (\(r_{MCC}=0.32\)). In Experiment 1, performances
were evaluated over 871 single-word concept tags used as queries (710
for non-experts and 201 for experts), and in Experiment 2, performances
were evaluated over 588 multi-word expressions.

\section{Results}\label{sec:results}

\subsection{Experiment 1}\label{experiment-1-1}

\hypertarget{tbl:exp1}{}
\begin{longtable}[]{@{}lll@{}}
\caption{\label{tbl:exp1}Concept detection performance on single-word
queries. Best scores in MCC and precision are emphasized in bold
}\tabularnewline
\toprule
& \textbf{Non-experts MCC} & \textbf{Experts MCC}\tabularnewline
\midrule
\endfirsthead
\toprule
& \textbf{Non-experts MCC} & \textbf{Experts MCC}\tabularnewline
\midrule
\endhead
\textbf{Keyword} & 0.14 & 0.14\tabularnewline
\textbf{LDA-Top30} & 0.37 & 0.34\tabularnewline
\textbf{Gibbs LDA} & 0.22 & 0.23\tabularnewline
\textbf{wAcAd} & 0.31 & 0.30\tabularnewline
\textbf{wAcAzAcAd} & 0.08 & 0.12\tabularnewline
\textbf{wAcAzAd} & 0.38 & 0.41\tabularnewline
\textbf{wAzAcAd} & 0.10 & 0.15\tabularnewline
\textbf{wAzAd} & 0.45 & \textbf{0.48}\tabularnewline
\textbf{wEcAd} & 0.14 & 0.18\tabularnewline
\textbf{wEcAzAcAd} & 0.12 & 0.16\tabularnewline
\textbf{wEcAzAd} & \textbf{0.46} & 0.45\tabularnewline
\bottomrule
\end{longtable}

As table~\ref{tbl:exp1} illustrates, the ``wAzAd'' and ``wEcAzAd''
heuristics were the most performant of all, outperforming the leading
LDA-Top30 method from Chartrand, Cheung, and Bouguessa (2017) by scores
ranging from 0,07 to 0,14 against non-expert and expert annotations.

Other trends can also be observed. Firstly, chains ending in ``cAd'' do
not fare well: their MCCs against non-expert annotations range from 0.10
to 0.31 (0.11 to 0.30 against expert annotations) while the chains
ending with ``zAd'' have MCCs ranging from 0.38 to 0.46 (0.41 to 0.48
against expert annotations). In fact, only and all zAd-ending chains
consistently outperformed LDA and keyword methods. Secondly, shorter
chains tend to have better performance than longer chains
(\(r=-0.48, p<0.001\) againts non-experts, \(r=-0.46, p<0.001\) against
experts).

\subsection{Experiment 2}\label{experiment-2-1}

\hypertarget{tbl:exp2}{}
\begin{longtable}[]{@{}lll@{}}
\caption{\label{tbl:exp2}Concept detection performance on compound word
queries }\tabularnewline
\toprule
& \textbf{Non-experts MCC} & \textbf{Experts MCC}\tabularnewline
\midrule
\endfirsthead
\toprule
& \textbf{Non-experts MCC} & \textbf{Experts MCC}\tabularnewline
\midrule
\endhead
\textbf{keyword} & 0,33 & 0,20\tabularnewline
\textbf{qEcAd} & 0,13 & 0,12\tabularnewline
\textbf{qEcAzAcAd} & 0,17 & 0,15\tabularnewline
\textbf{qEcAzAd} & \textbf{0,46} & \textbf{0,46}\tabularnewline
\bottomrule
\end{longtable}

As table~\ref{tbl:exp2} shows, the performances of the ``qEcAzAd'' hold
up with compound words, achieving similar MCCs as with single-word
queries. The same can probably be said for the ``qEcAd'' and
``qEcAzAcAd'' chains: while the former does worse against expert
annotations, and the latter does better against non-expert annotations,
it could only be indicative of noise in the data. The scores of the
keyword heuristic, on the other hand, have seen a significant uptick,
especially against non-expert annotations, where performance has more
than doubled.

\section{Discussion}\label{sec:discussion}

We made two claims about LCTM: (1) it makes for a better model of
higher-level entities like topics, which in turn translates in better
performances in concept detection and (2) it allows us to formulate
queries when concepts do not perfectly match with a word or word
expression in the corpus.

\subsection{Modelling and concept
detection}\label{modelling-and-concept-detection}

Concerning the first claim, results from Experiment 1 seem to validate
it, at least on the surface, as three of the heuristics managed to
provide us with better concept detection performance than what had
previously been achieved. These three chains also were more correlated
with experts and non-experts than experts and non-experts annotations
were correlated with each other.

Moreover, these results also suggest that the relationship between
concepts and the textual contexts in which they are present should not
be understood as a direct relation between words and concepts, but
rather, is mediated by higher-level entites like topics. Indeed, the
most successful chains are those that end by connecting those contexts
(in our case, the textual segments) to topics (i.e.~they end with a
``zAd'' operation). This confirms theoretical intuitions that we have
expressed elsewhere (Chartrand et al. 2016; Chartrand, Cheung, and
Bouguessa 2017; Chartrand no date).

Furthermore, it seems like it is the quality of the model that drives
the success of the LCTM in comparison with the LDA, because when the
same chain is used, LCTM does a lot better. The ``Gibbs LDA-Concrete
Assignment'' employs a ``wAzAd'' chain, but with a LDA model learned
using collapsed Gibbs sampling. Similarly, the LCTM model is learned
with an adapted collapsed Gibbs sampler. Therefore, the only difference
between those two methods lies in the model, and yet LCTM's ``wAzAd''
chain does more than twice as good as LDA's.

Other factors are also likely at play---in particular, chain length may
explain why some chains are better than others. For instance, some
chains seem to be achieving excessive recall (``wAcAzAcAd'', ``wAzAcAd''
and ``wEcAzAcAd'' in particular). This makes intuitive sense: the
``xAy'' operations all make it so that for every x, there can be more
than one y, as there usually is more than one token of x in the corpus,
and each token of x can be associated with a different type of y. As a
result, they end up overgenerating, and thus it is no wonder that they
would perform poorly in terms of the MCC. The relative success of
``wAcAd'' compared with ``wEcAd'' seems to come from the opposite excess
on the part of ``wEcAd'': given that the ``xEy'' operation only select
one y for every x, ``wEcAd'' only yields the segment assignations of
single LCTM concept, which makes for a very restricted concept
extension. On the other hand, with ``wAcAd'', individual words are
likely to be associated with various concepts. As a result, word queries
passing through the ``wAcAd'' chain yield the extension of several
concepts that are likely mobilized in the topics which mobilize the
queried concept---as such, they approximate the extension of a chain
that would use the topic extension like ``wAzAd''. For the same reason,
the ``xEy'' operation at the beginning of the ``wEcAzAcAd'' chain might
neutralize some of this long-chain effect, which would explain why it
does slightly better than ``wAcAzAcAd''.

\subsection{Concept detection of multiword
expressions}\label{concept-detection-of-multiword-expressions}

Concerning the second claim, it derives strong evidence from the success
of the ``qEcAzAd'' chain, which does as well on compound words as it did
on single words. This sustained performance may be somewhat surprising,
given that word embeddings composition is only an approximation of a
multi-word expression's meaning (e.g. Salehi, Cook, and Baldwin 2015).
However, single words themselves are often ambiguous (especially when
they are not chosen as research term, as is the case here); it is
possible that composition aliviates this ambiguity as to counter-balance
the imprecision it creates.

The relative success of the keyword heuristic on multiword expression
compared to single-word queries might also have to do with ambiguity. In
fact, most multiword expressions encountered among the annotation, like
``\emph{arbitre amiable compositeur}'' (amiable compositeur arbitrator)
and ``\emph{témoignage d'expert}'' (expert testimony) belong to the
technical juridical vocabulary. One can often find a precise definition
for it at the beginning of a law or a contract, or a detailed discussion
for its interpretation in the doctrine. Because jurists need to mitigate
the risk of coming to different interpretations of the same words, it is
perhaps more important than elsewhere to have technical concepts that
are explicitly linked to a body of text that can be leveraged for
interpretation. As a result, jurists have developed an habit of crafting
expressions that can be linked to a concept as unambiguously as
possible, and which are usually embedded in a set of words that can
rarely be seen elsewhere. Not only are these concepts unambiguous, but
often, the corresponding concept, being very technical, is also hard to
mobilize without using the corresponding expression. The keyword
heuristic thus employs expressions that have been refined for better
precision and recall---hence its success.

\subsection{Limitations}\label{limitations}

One of the motivations for employing LCTM was that it seemed like
employing words as a stand-ins for concepts was too indirect a way to
identify topics linked to said concept. One might have assumed that
translating that query into a vector on a word embedding space would
yield better results---but as we saw, one of the leading chains
(``wAzAd'') doesn't even leverage these representations. This might be
because annotators themselves were determining concept presence from a
word rather than a more direct expression of a concept. A fair test for
determining the best way to formulate a concept query would likely
require that annotators be given the task to identify the presence of
concepts formulated in other ways than corresponding words or
expression.

Another issue is that while testing for multiword expressions might give
us a hint as to the capacity of our LCTM chains to detect concepts
obtained from composition, it is not the most straightforward test for
the success of concept detection. We can expect a conceptual analyst to
compose concept representations to disambiguate a concept (e.g.~MIND -
OPINION to get the concept of MIND without contexts where ``mind'' is
used to mean ``opinion'', like ``in my mind, \ldots{}'') or to add or
remove a dimension of interest to it (e.g.~MIND + REASONING to study the
mind as a reasoning tool). Using composition in such ways is very
different to approximating a multiword expression, as is done in
Experiment 2. While its success is a good omen, we need to replicate
these results with tasks that are more in line with what conceptual
analysts are really likely to do.

On the more technical side, the relative success of online variational
Bayes compared to collapsed Gibbs sampling (which had already been
established by Chartrand, Cheung, and Bouguessa 2017) suggests that LCTM
might do even better with a different learning method. As such, it would
likely be worthwhile to adapt learning online variational Bayes
(Hoffman, Bach, and Blei 2010) or hybrid variational/Gibbs sampling
inference (Welling, Teh, and Kappen 2012) to the LCTM model in order to
learn better models.

\section{Conclusion}\label{sec:conclusion}

This paper sought to improve on existing concept detection methods by
modelling topics in a more theorically appropriate way as constituted of
concepts, and by enabling queries formulated in terms of coordinates on
the word embedding space. It pursued this objective by constructing
processing chains using LCTM models infered from a court decision corpus
using the method described by Hu and Tsujii (2016), and evaluated their
performance against annotations by legal experts and lay people.

It was successful on both counts. On single-word queries, some of the
chains achieved higher performance than the previous leading method, and
for reasons that seem to be due to the nature of the LCTM model. Queries
formulated as compositions of word embeddings were also tested as
approximation of multiword expressions and achieved equally high
results, demonstrating that our method can also successfully be used
with queries formulated as coordinates on the word embedding space.

\section{Acknowledgment}\label{acknowledgment}

This research was supported by the Social Sciences and Humanities
Research Council of Canada and enabled in part by the support provided
by WestGrid (\url{http://www.westgrid.ca}) and Compute Canada
(\url{www.computecanada.ca}). The author also wishes to thank Marc
Queudot, Mohamed Bouguessa and Jackie C.K. Cheung for their help and
comments.

\section*{References}\label{bibliography}
\addcontentsline{toc}{section}{References}

\hypertarget{refs}{}
\hypertarget{ref-Andow2016}{}
Andow, James. 2016. ``Qualitative tools and experimental philosophy.''
\emph{Philosophical Psychology} 29 (8):1128--41.

\hypertarget{ref-BaroniEtAl2014}{}
Baroni, Marco, Georgiana Dinu, and Germán Kruszewski. 2014. ``Don't
count, predict! A systematic comparison of context-counting vs.
context-predicting semantic vectors.'' In \emph{Proceedings of the 52nd
Annual Meeting of the Association for Computational Linguistics (Volume
1: Long Papers)}, 1:238--47.

\hypertarget{ref-BatmanghelichEtAl2016}{}
Batmanghelich, Kayhan, Ardavan Saeedi, Karthik Narasimhan, and Sam
Gershman. 2016. ``Nonparametric spherical topic modeling with word
embeddings.'' \emph{arXiv preprint arXiv:1604.00126}.

\hypertarget{ref-sep-analysis}{}
Beaney, Michael. 2018. ``Analysis.'' In \emph{The Stanford Encyclopedia
of Philosophy}, edited by Edward N. Zalta, Summer 2018. Metaphysics
Research Lab, Stanford University.

\hypertarget{ref-Bengio2003}{}
Bengio, Yoshua, Réjean Ducharme, Pascal Vincent, and Christian Jauvin.
2003. ``A neural probabilistic language model.'' \emph{Journal of
machine learning research} 3 (Feb):1137--55.

\hypertarget{ref-Blei2003}{}
Blei, David M., Andrew Y. Ng, and Michael I. Jordan. 2003. ``Latent
Dirichlet Allocation.'' \emph{Journal of machine Learning research} 3
(Jan):993--1022.

\hypertarget{ref-Blondel2010}{}
Blondel, Mathieu. 2010. ``Latent Dirichlet Allocation in Python.''
\emph{Mathieu's log}.
\href{\%7Bhttp://www.mblondel.org/journal/2010/08/21/latent-dirichlet-allocation-in-python/\%7D}{\{http://www.mblondel.org/journal/2010/08/21/latent-dirichlet-allocation-in-python/\}}.

\hypertarget{ref-Bluhm2013}{}
Bluhm, Roland. 2013. ``Don't Ask, Look! Linguistic Corpora as a Tool for
Conceptual Analysis.'' In \emph{Was dürfen wir glauben?: Was sollen wir
tun? Sektionsbeiträge des achten internationalen Kongresses der
Gesellschaft für Analytische Philosophie e.V.}, edited by Migue Hoeltje,
Thomas Spitzley, and Wolfgang Spohn, 7--15. DuEPublico.

\hypertarget{ref-BunkKrestel2018}{}
Bunk, Stefan, and Ralf Krestel. 2018. ``WELDA: Enhancing Topic Models by
Incorporating Local Word Context.'' In \emph{Proceedings of the 18th
ACM/IEEE on Joint Conference on Digital Libraries}, 293--302.

\hypertarget{ref-Carnap1950}{}
Carnap, Rudolf. 1950. \emph{Logical Foundations of Probability}.
Chicago: University of Chicago Press.

\hypertarget{ref-Chartrand2017artichaut}{}
Chartrand, Louis. 2017. ``La Philosophie Entre Intuition Et Empirie:
Comment les Études du Texte Peuvent Contribuer À Renouveler la Réflexion
Philosophique.'' \emph{Artichaud Magazine} 2017 (8 juin).

\hypertarget{ref-Chp1}{}
---------. no date. ``Similarity in conceptual analysis and concept as
proper function.''

\hypertarget{ref-Chartrand2017}{}
Chartrand, Louis, Jackie C. K. Cheung, and Mohamed Bouguessa. 2017.
``Detecting Large Concept Extensions for Conceptual Analysis.'' In
\emph{Machine Learning and Data Mining in Pattern Recognition}, 78--90.
Lecture Notes in Computer Science. Springer, Cham.

\hypertarget{ref-Chartrand2016}{}
Chartrand, Louis, Jean-Guy Meunier, Davide Pulizzotto, José López
González, Jean-François Chartier, Ngoc Tan Le, Francis Lareau, and
Julian Trujillo Amaya. 2016. ``CoFiH: A heuristic for concept discovery
in computer-assisted conceptual analysis.'' In \emph{JADT 2016 : 13ème
Journées internationales d'Analyse statistique des Données Textuelles}.
Vol. 1. Nice, France.

\hypertarget{ref-Collobert2008}{}
Collobert, Ronan, and Jason Weston. 2008. ``A unified architecture for
natural language processing: Deep neural networks with multitask
learning.'' In \emph{Proceedings of the 25th international conference on
Machine learning}, 160--67.

\hypertarget{ref-DasEtAl2015}{}
Das, Rajarshi, Manzil Zaheer, and Chris Dyer. 2015. ``Gaussian lda for
topic models with word embeddings.'' In \emph{Proceedings of the 53rd
Annual Meeting of the Association for Computational Linguistics and the
7th International Joint Conference on Natural Language Processing},
1:795--804.

\hypertarget{ref-Deerwester1990}{}
Deerwester, Scott, Susan T. Dumais, George W. Furnas, Thomas K.
Landauer, and Richard Harshman. 1990. ``Indexing by latent semantic
analysis.'' \emph{Journal of the American society for information
science} 41 (6):391--407.

\hypertarget{ref-ElAriniEtAl2012}{}
El-Arini, Khalid, Emily B. Fox, and Carlos Guestrin. 2012. ``Concept
modeling with superwords.'' \emph{arXiv preprint arXiv:1204.2523}.

\hypertarget{ref-Gabrilovich2007}{}
Gabrilovich, Evgeniy, and Shaul Markovitch. 2007. ``Computing Semantic
Relatedness Using Wikipedia-based Explicit Semantic Analysis.'' In
\emph{IJCAI 2007, Proceedings of the 20th International Joint Conference
on Artificial Intelligence, Hyderabad, India, January 6-12, 2007},
1606--11.

\hypertarget{ref-Gottron2011}{}
Gottron, Thomas, Maik Anderka, and Benno Stein. 2011. ``Insights into
explicit semantic analysis.'' In \emph{Proceedings of the 20th ACM
international conference on Information and knowledge management},
1961--4.

\hypertarget{ref-GriffithsSteyvers2004}{}
Griffiths, Thomas L., and Mark Steyvers. 2004. ``Finding scientific
topics.'' \emph{Proceedings of the National academy of Sciences} 101
(suppl 1):5228--35.

\hypertarget{ref-Haslanger2012}{}
Haslanger, Sally. 2012. \emph{Resisting Reality: Social Construction and
Social Critique}. Oxford: Oxford University Press.

\hypertarget{ref-Hoffman2010}{}
Hoffman, Matthew, Francis R. Bach, and David M. Blei. 2010. ``Online
learning for latent dirichlet allocation.'' In \emph{advances in neural
information processing systems}, 856--64.

\hypertarget{ref-Hofmann1999}{}
Hofmann, Thomas. 1999. ``Probabilistic latent semantic analysis.'' In
\emph{Proceedings of the Fifteenth conference on Uncertainty in
artificial intelligence}, 289--96.

\hypertarget{ref-HuTsujii2016}{}
Hu, Weihua, and Jun'ichi Tsujii. 2016. ``A latent concept topic model
for robust topic inference using word embeddings.'' In \emph{Proceedings
of the 54th Annual Meeting of the Association for Computational
Linguistics (Volume 2: Short Papers)}, 2:380--86.

\hypertarget{ref-LawEtAl2015}{}
Law, Jarvan, Hankz Hankui Zhuo, Junhua He, and Erhu Rong. 2017. ``LTSG:
Latent Topical Skip-Gram for Mutually Learning Topic Model and Vector
Representations.'' \emph{CoRR} abs/1702.07117.

\hypertarget{ref-LeEtAl2017}{}
Le, Tuan M. V., and Hady W. Lauw. 2017. ``Semantic Visualization for
Short Texts with Word Embeddings.'' In \emph{Proceedings of the
Twenty-Sixth International Joint Conference on Artificial Intelligence,
IJCAI-17}, 2074--80.

\hypertarget{ref-LiEtAl2016GPUDMM}{}
Li, Chenliang, Haoran Wang, Zhiqian Zhang, Aixin Sun, and Zongyang Ma.
2016. ``Topic modeling for short texts with auxiliary word embeddings.''
In \emph{Proceedings of the 39th International ACM SIGIR conference on
Research and Development in Information Retrieval}, 165--74.

\hypertarget{ref-LiEtAl2016TopicVec}{}
Li, Shaohua, Tat-Seng Chua, Jun Zhu, and Chunyan Miao. 2016.
``Generative topic embedding: a continuous representation of
documents.'' In \emph{Proceedings of the 54th Annual Meeting of the
Association for Computational Linguistics}, 1:666--75.

\hypertarget{ref-LiEtAl2016}{}
Li, Ximing, Jinjin Chi, Changchun Li, Jihong Ouyang, and Bo Fu. 2016.
``Integrating topic modeling with word embeddings by mixtures of vMFs.''
In \emph{Proceedings of COLING 2016, the 26th International Conference
on Computational Linguistics: Technical Papers}, 151--60.

\hypertarget{ref-LiEtAl2018}{}
Li, Ximing, Ang Zhang, Changchun Li, Lantian Guo, Wenting Wang, and
Jihong Ouyang. 2018. ``Relational Biterm Topic Model: Short-Text Topic
Modeling using Word Embeddings.'' \emph{The Computer Journal}.

\hypertarget{ref-LiuEtAl2015}{}
Liu, Yang, Zhiyuan Liu, Tat-Seng Chua, and Maosong Sun. 2015. ``Topical
Word Embeddings.'' In \emph{AAAI'15 Proceedings of the Twenty-Ninth AAAI
Conference on Artificial Intelligence}, 2418--24.

\hypertarget{ref-Meunier2005}{}
Meunier, Jean Guy, Ismail Biskri, and Dominic Forest. 2005.
``Classification and categorization in computer assisted reading and
analysis of texts.'' In \emph{Handbook of categorization in cognitive
science}, edited by Henri Cohen and Claire Lefebvre, 955--78. Elsevier.

\hypertarget{ref-Mikolov2013a}{}
Mikolov, Tomas, Kai Chen, Greg Corrado, and Jeffrey Dean. 2013.
``Efficient estimation of word representations in vector space.''
\emph{arXiv preprint arXiv:1301.3781}.

\hypertarget{ref-Mikolov2013b}{}
Mikolov, Tomas, Ilya Sutskever, Kai Chen, Greg S. Corrado, and Jeff
Dean. 2013. ``Distributed representations of words and phrases and their
compositionality.'' In \emph{Advances in neural information processing
systems}, 3111--9.

\hypertarget{ref-Moody2016}{}
Moody, Christopher E. 2016. ``Mixing Dirichlet Topic Models and Word
Embeddings to Make lda2vec.'' \emph{CoRR} abs/1605.02019.

\hypertarget{ref-Nguyen2015}{}
Nguyen, Dat Quoc, Richard Billingsley, Lan Du, and Mark Johnson. 2015.
``Improving topic models with latent feature word representations.''
\emph{Transactions of the Association for Computational Linguistics}
3:299--313.

\hypertarget{ref-PengEtAl2018}{}
Peng, Min, Qianqian Xie, Yanchun Zhang, Hua Wang, Xiuzhen Zhang, Jimin
Huang, and Gang Tian. 2018. ``Neural Sparse Topical Coding.'' In
\emph{Proceedings of the 56th Annual Meeting of the Association for
Computational Linguistics (Volume 1: Long Papers)}, 2332--40. Melbourne,
Australia: Association for Computational Linguistics.

\hypertarget{ref-Pennington2014}{}
Pennington, Jeffrey, Richard Socher, and Christopher Manning. 2014.
``Glove: Global vectors for word representation.'' In \emph{Proceedings
of the 2014 conference on empirical methods in natural language
processing (EMNLP)}, 1532--43.

\hypertarget{ref-PotapenkoEtAl2017}{}
Potapenko, Anna, Artem Popov, and Konstantin Vorontsov. 2017.
``Interpretable probabilistic embeddings: bridging the gap between topic
models and neural networks.'' In \emph{Conference on Artificial
Intelligence and Natural Language}, 167--80.

\hypertarget{ref-Pulizzotto2016}{}
Pulizzotto, Davide, José A. Lopez, Jean-François Chartier Jean-Guy,
Meunier1 Louis Chartrand1 Francis Lareau Tan, and Le Ngoc. 2016.
``Recherche de «périsegments» dans un contexte d'analyse conceptuelle
assistée par ordinateur: le concept d'«esprit» chez Peirce.'' In
\emph{JEP-TALN-RECITAL 2016}. Vol. 2. Paris.

\hypertarget{ref-Pust2000}{}
Pust, Joel. 2000. \emph{Intuitions as Evidence}. New York: Routledge.

\hypertarget{ref-Sahlgren2008}{}
Sahlgren, Magnus. 2008. ``The distributional hypothesis.'' \emph{Italian
Journal of Disability Studies} 20:33--53.

\hypertarget{ref-SalehiEtAl2015}{}
Salehi, Bahar, Paul Cook, and Timothy Baldwin. 2015. ``A word embedding
approach to predicting the compositionality of multiword expressions.''
In \emph{Proceedings of the 2015 Conference of the North American
Chapter of the Association for Computational Linguistics: Human Language
Technologies}, 977--83.

\hypertarget{ref-TangEtAl2018}{}
Tang, Yi-Kun, Xian-Ling Mao, Heyan Huang, Xuewen Shi, and Guihua Wen.
2018. ``Conceptualization topic modeling.'' \emph{Multimedia Tools and
Applications} 77 (3):3455--71.

\hypertarget{ref-WangEtAl2017}{}
Wang, Bo, Maria Liakata, Arkaitz Zubiaga, and Rob Procter. 2017. ``A
Hierarchical Topic Modelling Approach for Tweet Clustering.'' In
\emph{International Conference on Social Informatics}, 378--90.

\hypertarget{ref-WellingEtAl2012}{}
Welling, Max, Yee Whye Teh, and Hilbert Kappen. 2012. ``Hybrid
variational/Gibbs collapsed inference in topic models.'' \emph{arXiv
preprint arXiv:1206.3297}.

\hypertarget{ref-XunEtAl2017}{}
Xun, Guangxu, Yaliang Li, Wayne Xin Zhao, Jing Gao, and Aidong Zhang.
2017. ``A correlated topic model using word embeddings.'' In
\emph{Proceedings of the 26th International Joint Conference on
Artificial Intelligence}, 4207--13.

\hypertarget{ref-ZhangEtAl2019}{}
Zhang, Xianchao, Ran Feng, and Wenxin Liang. 2019. ``Short Text Topic
Model with Word Embeddings and Context Information.'' In \emph{Recent
Advances in Information and Communication Technology 2018}, edited by
Herwig Unger, Sunantha Sodsee, and Phayung Meesad, 55--64. Cham:
Springer International Publishing.

\hypertarget{ref-ZhaoEtAl2018}{}
Zhao, He, Lan Du, Wray Buntine, and Mingyuan Zhou. 2018. ``Inter and
Intra Topic Structure Learning with Word Embeddings.'' In
\emph{International Conference on Machine Learning}, 5887--96.

\end{document}